\documentclass[letterpaper]{article} 
\usepackage{aaai2026}
\usepackage{times}  
\usepackage{helvet}  
\usepackage{courier}  
\usepackage[hyphens]{url}  
\usepackage{graphicx} 
\usepackage{natbib}  
\usepackage{caption} 
\usepackage{algorithm}
\usepackage{algorithmic}
\usepackage{amsmath, amssymb}
\usepackage{tabularx}
\usepackage{booktabs}
\usepackage{pifont} 
\usepackage{graphicx}
\usepackage{multirow}
\usepackage{adjustbox}
\usepackage{makecell}
\usepackage{tikz}
\usepackage[utf8]{inputenc}
\usepackage{newunicodechar}
\newunicodechar{①}{1}
\newunicodechar{②}{2}
\newunicodechar{③}{3}
\newunicodechar{④}{4}

\usepackage{newfloat}
\usepackage{listings}
\DeclareCaptionStyle{ruled}{labelfont=normalfont,labelsep=colon,strut=off} 
\floatstyle{ruled}
\newfloat{listing}{tb}{lst}{}
\floatname{listing}{Listing}
\title{$\Delta$-NeRF: Incremental Refinement of Neural Radiance Fields through Residual Control and Knowledge Transfer}
\author {
    Kriti Ghosh \textsuperscript{\rm 1},
    Devjyoti Chakraborty \textsuperscript{\rm 1},
    Lakshmish Ramaswamy \textsuperscript{\rm 1,2},
    Suchendra M. Bhandarkar
    \textsuperscript{\rm 1,2},
    In Kee Kim \textsuperscript{\rm 1,2},
    Nancy O'Hare \textsuperscript{\rm 3},
    Deepak Mishra \textsuperscript{\rm 2,3}
}
\affiliations {
    \textsuperscript{\rm 1}School of Computing\\
    \textsuperscript{\rm 2}Institute for Artificial Intelligence\\
    \textsuperscript{\rm 3}Center for Geospatial
Research, Dept. of Geography\\
The University of Georgia, Athens, Georgia 30602, USA
}
\usepackage{bibentry}

\begin{document}

\maketitle

\begin{abstract}

Neural Radiance Fields (NeRFs) have demonstrated remarkable capabilities in 3D reconstruction and novel view synthesis. However, most existing NeRF frameworks require complete retraining when new views are introduced incrementally, limiting their applicability in domains where data arrives sequentially. This limitation is particularly problematic in satellite-based terrain analysis, where regions are repeatedly observed over time. Incremental refinement of NeRFs remains underexplored, and naive approaches suffer from catastrophic forgetting when past data is unavailable. We propose $\Delta$-NeRF, a unique modular residual framework for incremental NeRF refinement. $\Delta$-NeRF introduces several novel techniques including: (1) a residual controller that injects per-layer corrections into a frozen base NeRF, enabling refinement without access to past data; (2) an uncertainty-aware gating mechanism that prevents overcorrection by adaptively combining base and refined predictions; and (3) a view selection strategy that reduces training data by up to 47\% while maintaining performance. Additionally, we employ knowledge distillation to compress the enhanced model into a compact student network (20\% of original size). Experiments on satellite imagery demonstrate that $\Delta$-NeRF achieves performance comparable to joint training while reducing training time by 30-42\%. $\Delta$-NeRF consistently outperforms existing baselines, achieving an improvement of up to 43.5\% in PSNR over naive fine-tuning and surpassing joint training on some metrics.

\end{abstract}

\section{Introduction}

Neural Radiance Fields (NeRFs) have emerged as a powerful technique for photorealistic novel view synthesis and 3D reconstruction~\cite{nerf}.  
Recent advances have extended NeRFs to 
diverse domains including fast training and rendering~\cite{plenoxels,ngp}, editable view synthesis~\cite{edit}, 
reflectance-aware modeling and relighting using Bidirectional Reflectance Distribution Function (BRDF) decomposition~\cite{nero,nerd}, and remote sensing and aerial scene reconstruction~\cite{satnerf,sps}.
However, existing NeRF methods assume all training views are {\em available upfront}.
This assumption is overly restrictive in scenarios where the data (i.e., views) are collected incrementally over time. This critical limitation impairs the effectiveness and practical viability of these models across many real-world application domains. 
For instance, in satellite-based terrain modeling, it is common to have a partial set of views at the beginning with additional views being acquired over time ~\cite{season} . 

To the best of our knowledge, current state-of-the-art (SOTA) methods lack a robust and computationally efficient mechanism for integrating additional views acquired post hoc into a trained NeRF model. Training the model from scratch for integrating new views is highly redundant and expensive. The primary objective of this paper is to address this critical need for \textit{incremental NeRF refinement}. We design robust and efficient mechanisms for incorporating additional views into pre-trained NeRF models without requiring full retraining. While our work is inspired by the satellite remote sensing of terrain analysis domain, the challenges we address in this work are generic to many NeRF applications. We note that incremental learning remains underexplored for NeRFs in general and for satellite imagery in particular. 

A major challenge in incremental NeRF refinement is \textit{catastrophic forgetting}~\cite{forget,clnerf2,ilnerf}, wherein naive fine-tuning of a pretrained NeRF model on newer views often leads to overfitting to this new data thus causing performance degradation~\cite{ilnerf}. Most existing approaches to mitigate catastrophic forgetting in problem domains such as incremental classification, object detection and 3D scene reconstruction involve regularization techniques such as Elastic Weight Consolidation (EWC) or memory optimization mechanisms such as Knowledge Distillation (KD) with memory replay or generative replay~\cite{od,clnerf2, ilnerf}. Both EWC and KD have certain inherent limitations; while EWC suffers from rigidity and adaptation issues, KD requires access to representative data from the past, which may not always be available. Furthermore, the effectiveness of these techniques in the context of NeRFs is not well studied. 

The inherent nature of satellite imagery introduces various additional challenges to incremental refinement, that are unique to Earth observation data. Due to the limited viewing angles covered by satellite images, the reconstruction problem becomes more challenging. Moreover, the variability of appearance due to illumination changes (BRDF effects), shadows, occlusions, changes in atmospheric and seasonal conditions, and urban development makes it difficult for the model to adapt incrementally without significant degradation in generalization capability.  

We present $\Delta$-NeRF, a plug-and-play framework that incrementally adapts to new satellite views by learning only the residual corrections, to modify a frozen base model without the need for explicit memory replay. Inspired by ControlNet~\cite{controlnet} and residual learning, $\Delta $-NeRF offers a scalable, efficient, and modular framework that learns incrementally from newer views without sacrificing prior knowledge. Unlike some prior works that focus on continual scene edits on indoor or outdoor scenes, we focus on incremental refinement to enhance the quality of NeRF models by integrating post-hoc views into pretrained models. This modular setup improves interpretability, allows distillation of the updated model into a compact student model, and avoids the computational overhead of full retraining.

Our research contributions include:
\begin{itemize}
    
    \item We design a novel modular zero-initialized residual network controller that injects per-layer feature residuals into a frozen base NeRF for enabling incremental corrections and avoiding catastrophic forgetting by design. This permits incremental training without requiring access to past data or any form of memory replay. 
    \item We incorporate a novel error-and-uncertainty-aware gating mechanism that combines the previous and newly learned corrections to obtain the best RGB prediction.
    \item We employ a two-stage pipeline optimization: a depth-aware view selection method to maximize information diversity while reducing training cost, and knowledge distillation to compress the controller-enhanced model into a compact student ($\sim$20\% of original size) for efficient deployment.
    
    \item We show that, after incremental training, the performance of the $\Delta$-NeRF framework improves beyond that of the initial model and matches the all-at-once jointly trained performance. $\Delta$-NeRF  displays performance comparable with existing baselines. 

\end{itemize}

Experiments show that the incrementally trained $\Delta$-NeRF model achieves performance levels comparable to baselines while reducing the computational overhead by up to 35–45\%.

\section{Background and Motivation}

\paragraph{Neural Radiance Field (NeRF):}
Given, a 3D location $\mathbf{x} \in \mathbb{R}^3$ and viewing direction $\mathbf{d} \in \mathbb{R}^3$ the NeRF learns to synthesize novel views by predicting color, $\mathbf{c} \in [0,1]^3$ and volume density $\sigma \in \mathbb{R}_+$ for points sampled along camera rays, and composing them to render realistic images. The continuous NeRF rendering computes the final color of a ray $r(t) = \mathbf{o} + t\mathbf{d}$ by integrating emitted color along the ray, weighted by volume density and accumulated transmittance:

\begin{equation}
F(\mathbf{x}, \mathbf{d}) \rightarrow (\mathbf{c}, \sigma), \quad 
\mathbf{c}(r) = \int_{t_n}^{t_f} T(t) \, \sigma(t) \, \mathbf{c}(t) \, dt
\label{eq:init-eqn}
\end{equation}
Here,$\mathbf{o}$ and $\mathbf{d}$ are the origin and direction of the ray; $t \in [t_n, t_f]$ is the distance along the ray, from near bound $t_n$ to far bound $t_f$; $T(t)$ is the accumulated transmittance (i.e., the probability that light reaches $t$ without hitting another particle); $\sigma(t)$ is the predicted density; and $\mathbf{c}(t)$ is the emitted RGB color at that point. However, NeRF struggles under complex illumination conditions and limited view availability, as are commonly observed in satellite imagery. 

There are a few NeRF variants which try to address these limitations. \textbf{Shadow-NeRF (S-NeRF)}~\cite{snerf} addresses this by modeling sun visibility and diffuse skylight to better predict appearance. \textbf{Satellite-NeRF (Sat-NeRF)}~\cite{satnerf}   extends S-NeRF with a Rational Polynomial Camera (RPC) model and uncertainty-based handling of transient objects. \textbf{SparseSat-NeRF (SpS-NeRF)}~\cite{sps} adapts Sat-NeRF to sparse views using depth priors.

\paragraph{Need for Incremental Refinement:} 

In the naive approach, NeRF models must be re-trained from scratch on all views, old and new, when new data arrive. For instance ( Table~\ref{tab:training_time}), training on 7 satellite views takes 4–5 hours, and adding 10 more images later requires an additional 10 hours, even if prior data remain unchanged. These times are measured using Sat-NeRF on an NVIDIA RTX A6000 GPU with authors' default configuration. While effective, this is inefficient for continual updates. Naive fine-tuning reduces redundancy but causes catastrophic forgetting. In Earth observation, where regions are frequently revisited, scalable refinement is crucial. We propose an incremental framework that avoids full retraining by selecting a compact, informative subset of views, reducing both training time and data.
\begin{table}[h]
\centering
\setlength{\tabcolsep}{2pt} 
\begin{tabular}{ccccc}
\toprule
\makecell{Initial \\ Views} & \makecell{Additional \\ Views} & \makecell{Initial \\ Time (hrs.)} & \makecell{Retraining \\ Time (hrs.)} & \makecell{Total \\ Time (hrs.)} \\
\midrule
7 & 10 & $\sim$4 & $\sim$10 & $\sim$14 \\
\bottomrule
\end{tabular}
\caption{Training time in full retraining}
\label{tab:training_time}
\end{table}

\paragraph{State of the Art and Limitations:}

Most NeRF variants assume full data access upfront, making them poorly suited for scenarios where new data arrive over time. Naive fine-tuning causes forgetting often requiring retraining. Some recent work aims to address this issue. CLNeRF~\cite{clnerf2} uses generative replay with Instant Neural Graphics Primitives (Instant-NGPs) to prevent forgetting without storing prior images, but struggles with transient elements due to overfitting by the NGPs. CL-NeRF~\cite{clnerf1} introduces an expert adapter and conflict-aware KD with a soft gating mask, but is evaluated only on two synthetic and two indoor-outdoor datasets, not on satellite imagery. IL-NeRF~\cite{ilnerf} is a replay-based method that computes new camera poses based on past camera poses, but is less effective for large-scale scenes with limited overlap. Other existing methods~\cite{meil,incr1,instant} use different forms of ray replay, regularization, and replay buffer mechanisms to address the incremental learning (IL) problem in NeRF. However, access to past data may not always be feasible.

In contrast, our method introduces a residual modulation controller inspired by ControlNet~\cite{controlnet}, which learns to modulate internal layer-wise features. Also, unlike prior works, our focus is on incremental learning for large-scale satellite imagery. Moreover, we integrate uncertainty-aware gated inference with optimized view selection, offering robustness in sparse-view scenarios without catastrophic forgetting.

\paragraph{Research Challenges:}
The problem of incremental refinement of the existing model can be defined as follows.
Given a pre-trained NeRF model \( M_b \) trained on an initial set of satellite views $\mathcal{I}_{\text{base}} = \{ v_1, v_2, \ldots, v_N \}$, and a new set of views $\mathcal{I}_{\text{new}} = \{ v_{N+1}, v_{N+2}, \ldots, v_{N+n} \}$, our goal is to incrementally update the model to learn from $\mathcal{I}_{\text{new}}$ without degrading performance on $\mathcal{I}_{\text{base}}$. Crucially, this update must occur \textbf{without retraining from scratch}, \textbf{without accessing past data}, and \textbf{without significantly increasing the model size}. The above problem formulation presents the following core challenges:
\textbf{(C1)} Preventing catastrophic forgetting \textbf{(C2)} Learning new information available through the $\mathcal{I}_{\text{new}}$ \textbf{(C3)} Refining representations where prior performance is acceptable but suboptimal \textbf{(C4)} Combining old and new to generate a coherent representation of the entire scene from separate training sessions. Note that having no access to past data during incremental training limits conventional fine-tuning or replay-based approaches.

\section{$\Delta $-NeRF Framework}

\begin{figure}[t]
    \centering
    \includegraphics[width=0.9\columnwidth]{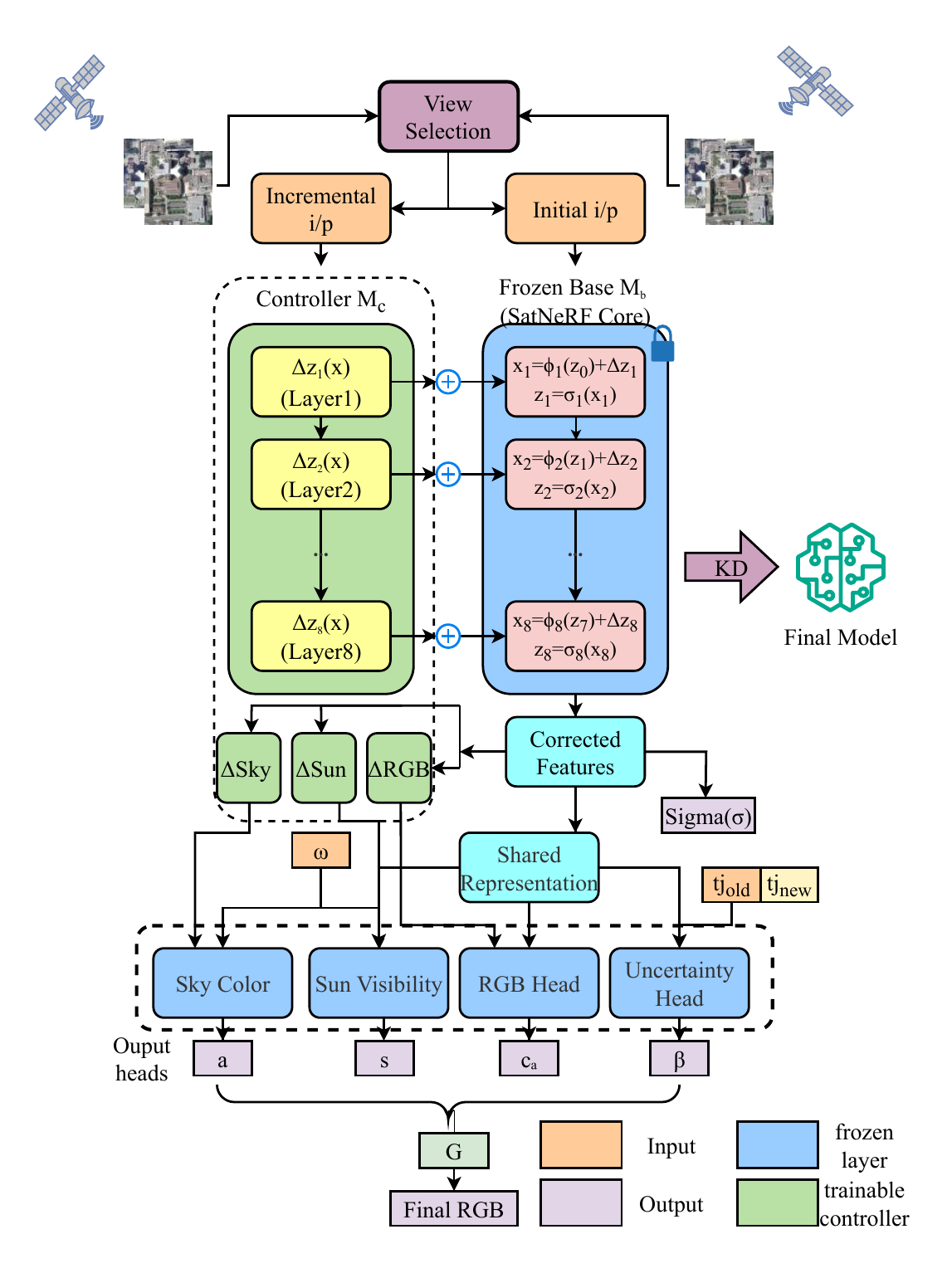}
    \caption{$\Delta $-NeRF architecture. The residual controller ($\Delta$Mc) injects feature-level residuals into a frozen base model (Mb). Appearance heads (RGB, sun visibility, sky color) are corrected using dedicated residual paths. Uncertainty-aware gating combines base and corrected outputs.}
    \label{fig:ctrl}
\end{figure}

\subsection{Overview}
To address the aforementioned challenges under the given constraints, we propose $\Delta$-NeRF, a \textbf{residual controller-based approach}, depicted in Figure~\ref{fig:ctrl}. In $\Delta$-NeRF, we handle the catastrophic forgetting problem in multiple stages. First, the controller learns corrections to $M_b$ based on $\mathcal{I}_{\text{new}}$ while $M_b$ remains frozen. 
This design \textbf{ensures the preservation of prior knowledge} (\textbf{C1}), \textbf{learns new information}  (\textbf{C2}), and \textbf{modifies only those features that require correction}, thus enabling incremental refinement (\textbf{C3}). Second, at inference time, we perform a \textbf{gated fusion of base and controller corrected outputs}, where ray-level error and uncertainty guide the combination (\textbf{C4}). This also ensures that the controller corrections do not overwrite the old model knowledge, a critical step in \textbf{preventing catastrophic forgetting}. Finally, the base and controller models are distilled into a \textbf{compact unified model} using KD that preserves both old and new knowledge.

\subsection{Residual Controller for Incremental Learning}

To preserve prior knowledge while adapting to novel views, we introduce a residual controller, inspired by ControlNet~\cite{controlnet}, to refine a base model trained on the initial dataset $\mathcal{I}_{\text{base}}$. When new data becomes available, we assume:

\[
\mathcal{I}_\text{base} \cap \mathcal{I}_\text{incr} = \emptyset,\quad
\text{scene}(\mathcal{I}_\text{base}) = \text{scene}(\mathcal{I}_\text{incr})
\]
So the scenes are overlapping. This models a realistic setting in satellite imagery, where the same region is incrementally observed from multiple passes with varying sun angles and viewing geometries. Now, instead of retraining or fine-tuning the base Sat-NeRF model \( M_b \), we freeze it and train a multilayer controller that learns intermediate feature corrections from new data $\mathcal{I}_{\text{new}}$. 
To ensure stability and preserve prior knowledge, we zero-initialize the residual controller. This guarantees that at the start of incremental training, the model's predictions exactly match the frozen base, and the controller only learns corrections where necessary. Such initialization acts as a strong regularizer, which is the first step toward preventing catastrophic forgetting  while enabling targeted adaptation to new data. 

Let  $\Delta M_c$ be the trainable residual controller. As displayed in Figure~\ref{fig:ctrl}, the controller mirrors the architecture of the base trunk $FC_b^{\text{trunk}}$. 
For each spatial location $x$, ray direction $\omega$, and timestamp $t_j$, the base trunk $FC_b^{\text{trunk}}$ computes intermediate shared features via an eight-layer fully-connected network where  $z = FC_b^{\text{trunk}}(x)$. The controller trunk $FC_c^{\text{trunk}}$ predicts residuals $\Delta z = \Delta FC_c^{\text{trunk}}(x)$, which are added layer-wise to obtain corrected shared features.

Given input locations $\mathbf{x}$, we first compute residual feature corrections layer-wise from the controller: \[\Delta \mathbf{z}_i(\mathbf{x}) = \psi_i\left( \mathbf{z}_{i-1}^{\text{ctrl}}(\mathbf{x}) \right), \quad i = 1, \dots, 8\] where $\psi_i$ denotes the $i$-th controller residual layer, and $\mathbf{z}_{i-1}^{\text{ctrl}}$ is the controller's intermediate feature. These residuals are then injected into the base trunk $f_b^{\text{trunk}}$ layer-by-layer: \[
\mathbf{z}_i(\mathbf{x}) = \sigma_i\left( \phi_i(\mathbf{z}_{i-1}(\mathbf{x})) + \Delta \mathbf{z}_i(\mathbf{x}) \right), \quad i = 1, \dots, 8 \]
where $\phi_i$ is the $i$-th layer of the frozen base trunk and $\sigma_i$ is the corresponding activation function. Finally, the corrected shared feature is: \[
\tilde{\mathbf{z}}(\mathbf{x}) = \mathbf{z}_8(\mathbf{x}).\] 

For final RGB prediction we further inject residuals to appearance related heads (\textit{RGB, sun visibility, sky color}). No residuals are applied to physical outputs such as \textit{density} or uncertainty. 
Given these corrected shared features, appearance prediction head receives this modulated input feature vector, and the controller supplies a corresponding residual correction. Layer-wise residual injection for any output head with $N$ layers (e.g., R\textit{GB, sun visibility, sky color}) that takes the corrected shared feature $\tilde{\mathbf{z}}$ and optionally some additional input $\mathbf{u}$ (e.g., \textit{sun direction}), the residual-modulated computation proceeds as:\[
\mathbf{h}_i =
\begin{cases}
f_1^{\text{head}}(\tilde{\mathbf{z}}, \mathbf{u}) + \Delta f_1^{\text{head}}(\tilde{\mathbf{z}}, \mathbf{u}), & i = 1 \\
f_i^{\text{head}}(\rho(\mathbf{h}_{i-1})) + \Delta f_i^{\text{head}}(\rho(\hat{\mathbf{h}}_{i-1})), & i = 2, \dots, N
\end{cases} \] where  $\rho(\cdot)$ is the activation function, $\mathbf{h}_{i-1}$ is the base model's previous layer output, $\hat{\mathbf{h}}_{i-1}$ is the controller’s previous layer output, and $\Delta f_i^{\text{head}}(\cdot)$ is the controller’s residual prediction.

The final output is given by:
\[
\hat{\mathbf{y}}(x) = \sigma(\mathbf{h}_N),
\]
where $\sigma(\cdot)$ is a head-specific output activation (e.g., sigmoid or ReLU). Note that the modular controller adapts to new views by modulating intermediate features, while the frozen base retains prior knowledge. The original Sat-NeRF rendering equation remains unchanged.

\subsection{Incremental Uncertainty Coefficient}

As with Sat-NeRF~\cite{satnerf}, we utilize an uncertainty coefficient \(\beta\). Specifically, \(\beta(\mathbf{x}, t_j)\) models the uncertainty at each 3D point \(\mathbf{x}\) given the image-specific time embedding \(t_j\), and is aggregated along each ray using transmittance \(T_i\) and opacity \(\alpha_i\) as:
\begin{equation}
\beta(r) = \sum_{i=1}^N T_i \alpha_i \beta(\mathbf{x}_i, t_j).
\end{equation}  
where \( \mathbf{x}_i \) is the \( i \)-th 3D point along ray \( r \), and \( \mathbf{t}_j \) is the learned embedding corresponding to the input image index \( j \) from which the ray originates. Each image has a unique embedding vector \( \mathbf{t}_j \), shared across all rays from that image.

In our incremental setup, we extend the embedding table to add new image indices while freezing the first \( j_\text{old} \) entries. For any new image \( j' \geq j_\text{old} \),a fresh embedding \( \mathbf{t}_{j'} \) is learned and used with the frozen base model to compute \( \beta(\mathbf{x}_i, \mathbf{t}_{j'}) \). As the scene remains unchanged, we rely on the base model’s generalization to estimate uncertainty for the new views.

\subsection{Uncertainty-aware Gated Inference}
To mitigate the issue of over-correction by the controller during incremental training, we introduce a soft gating mechanism that combines the base model's output with the residual-corrected output based on their relative reliability. Following Sat-NeRF's~\cite{satnerf} uncertainty modeling, we use the base model's predicted uncertainty \(\beta\) to define a confidence-based fusion gate. Given a ray \(r\), let \(\hat{\mathbf{c}}_{\text{base}}(r)\) and \(\hat{\mathbf{c}}_{\text{res}}(r)\) be the RGB predictions from the frozen base model and the residual-corrected student model respectively. We compute per-ray errors \(e_{\text{base}}, e_{\text{res}}\) with respect to the ground truth and define a confidence score \(s = \frac{1}{\beta(r)}\). The fusion gate is then given by:
\[
g(r) = \sigma\left(\lambda \cdot s \cdot \left( e_{\text{base}}(r) - e_{\text{res}}(r) \right)\right),
\]
where \(\sigma(\cdot)\) is the sigmoid function and \(\lambda\) is a scaling hyperparameter. The final RGB prediction is a weighted average:
\[
\hat{\mathbf{c}}_{\text{fused}}(r) = g(r) \cdot \hat{\mathbf{c}}_{\text{res}}(r) + (1 - g(r)) \cdot \hat{\mathbf{c}}_{\text{base}}(r).
\]
This formulation ensures that the base model output is properly combined with the controller corrected base model output to provide the best RGB prediction.  It enables more robust inference during incremental updates without sacrificing generalization to previously learned data.

\paragraph{Incremental Training Objective:}  
We supervise the model \( (\hat{c}, \hat{\sigma}) \) using the ground truth RGB values \( c_{\text{gt}} \). Following the Sat-NeRF~\cite{satnerf} formulation, the overall training loss includes an RGB loss component $\mathcal{L}_{\text{RGB}}$ given by:
\[
\mathcal{L}_{\text{RGB}} = \| \hat{c} - c_{\text{gt}} \|^2, \quad epoch= 0,1
\]  
\[
\mathcal{L}_{\text{RGB}} = \sum_{r} \frac{\|\mathbf{c}(r) - \mathbf{c}^{\text{GT}}(r)\|^2}{2\beta(r)^2} + \log \beta(r),  epoch >=2.
\label{eq:color_loss}
\]
Additionally, as a safeguard against forgetting, the overall training loss includes a KD loss component $\mathcal{L}_{\text{KD}}$ given by: 
\[
\mathcal{L}_{\text{KD}} = \| \hat{c} - c_{\text{base}} \|^2, \quad
\mathcal{L}_{\text{total}} = \mathcal{L}_{\text{RGB}} + \lambda \mathcal{L}_{\text{KD}}
\] 

The above formulation of $\mathcal{L}_{\text{total}}$ enables continual adaptation across new data without catastrophic forgetting.

\subsection{Two-pronged Optimization of Data and Model}

To address the computational overhead of NeRF training, we employ a pre-training optimized view selection and post-training KD phase. 
Both initial and incremental datasets are optimized using our view selection method, which preserves maximum information while reducing training time. 
After incremental training, KD produces a smaller more compact model that incorporates both old and new knowledge. 

\subsubsection{Data Optimization via Optimal View Selection:} Let $\mathcal{I} = \{ v_1, v_2, \ldots, v_N \}$ denote the set of satellite views to be optimized, each associated with RGB content, RPC metadata, sun angles, and a DSM tile. Our goal is to select a compact and informative subset $\mathcal{S} \subset \mathcal{I}$ for NeRF training.

We compute joint embeddings for each $v_i$ using: 
(i) appearance features $F_{\text{rgb}}(v_i)$ from a frozen ResNet18, 
(ii) DSM-based terrain features $F_{\text{dsm}}(I_i)$, and 
(iii) solar-time metadata $f_{\text{meta}}(v_i)$. These embeddings are normalized, reduced via PCA, and clustered to select diverse representatives close to the cluster centroids. 

To ensure terrain diversity, we estimate single-view depth maps $D_i$ using the MiDaS system~\cite{midas} and compute the coverage ratio \( C(\mathcal{S}) = \frac{\text{range}(\mathcal{S})}{\text{range}(\mathcal{I})} \).

We greedily expand $\mathcal{S}$ until a coverage threshold is met, then prune redundant views.
Residual views $v_j \notin \mathcal{S}$ are ranked by their silhouette score when added to $\mathcal{S}$, encouraging selection of complementary perspectives. A final validation step evaluates the representation completeness based on appearance, depth, metadata and representational features to ensure that $\mathcal{S}$ preserves the structural diversity of $\mathcal{I}$.

\subsubsection{Model Optimization via Knowledge Distillation:}

After the incremental training phase, we distill the enhanced model into a compact student network $M_{\text{student}}$. The student learns to mimic the behavior of the composite model using a distillation loss:

\[
{\mathcal{L}_{\text{KD}}^{\text{rgb}} = (\left\| \hat{\mathbf{c}}_{\text{student}}(x) - \hat{\mathbf{c}}_{\text{composite}}(x) \right\|_2)^2}.
\]
KD produces a lightweight model ($\sim$20\% of  original size) that preserves the performance while enabling deployment of resource-constrained platforms.

\section{Experiments and Results}
\begin{figure*}[t]
    \centering
    
     \includegraphics[width=\linewidth,height=5.5cm, keepaspectratio]{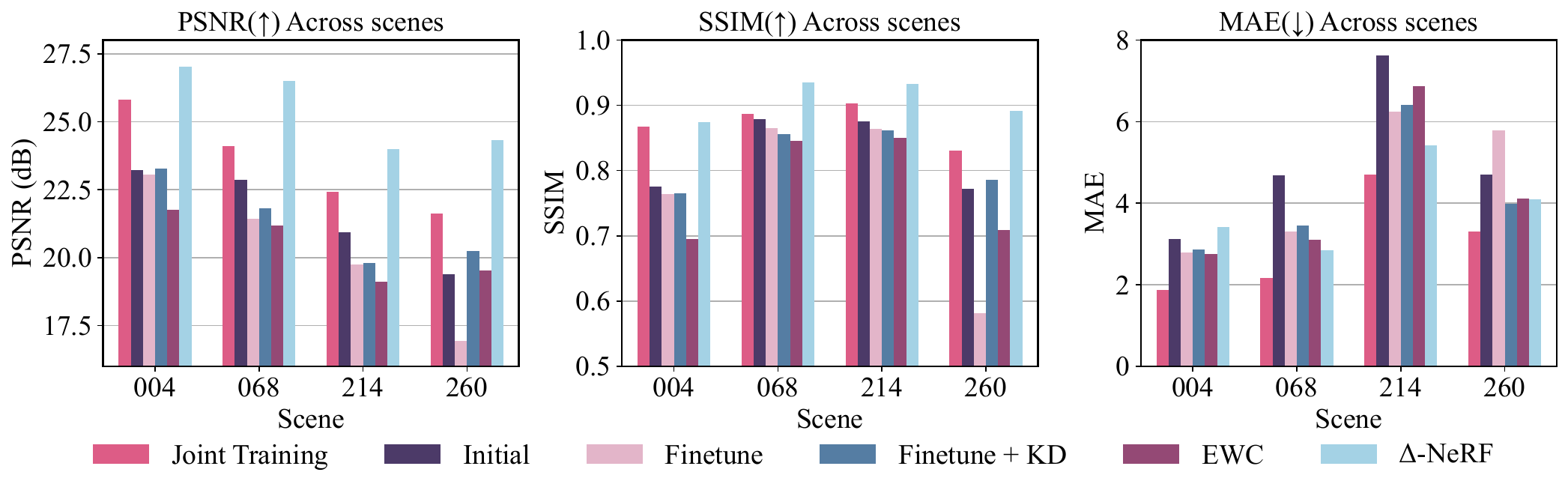}
    \caption{Performance comparison with baselines. Our $\Delta$-NeRF performs better than joint performance in PSNR and SSIM. $\Delta$-NeRF gives comparable performance with other baselines in MAE. MAE is not gated.}
    \label{fig:metrics_barplot}
\end{figure*}

\begin{figure*}[t]
  \centering
    
    \includegraphics[width=\linewidth,height=3cm, keepaspectratio]{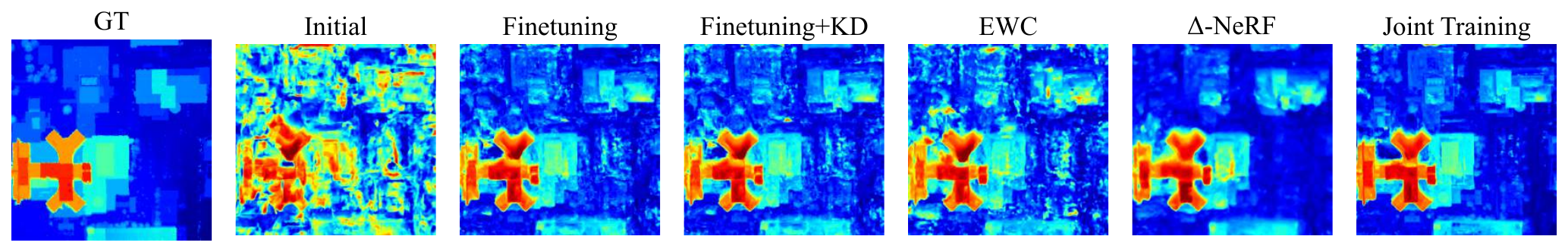}

  \caption{Qualitative comparisons of DSM reconstruction after incremental learning across baselines. Our incremental approach, $\Delta$-NeRF, significantly improves the noisy initial model, provides comparable qualitative performance with other baselines, and approach the performance of the joint training upper bound.}
  \label{fig:qual_rgb_dsm}
\end{figure*}

\begin{figure*}[t]
    \centering
    \includegraphics[width=\linewidth]{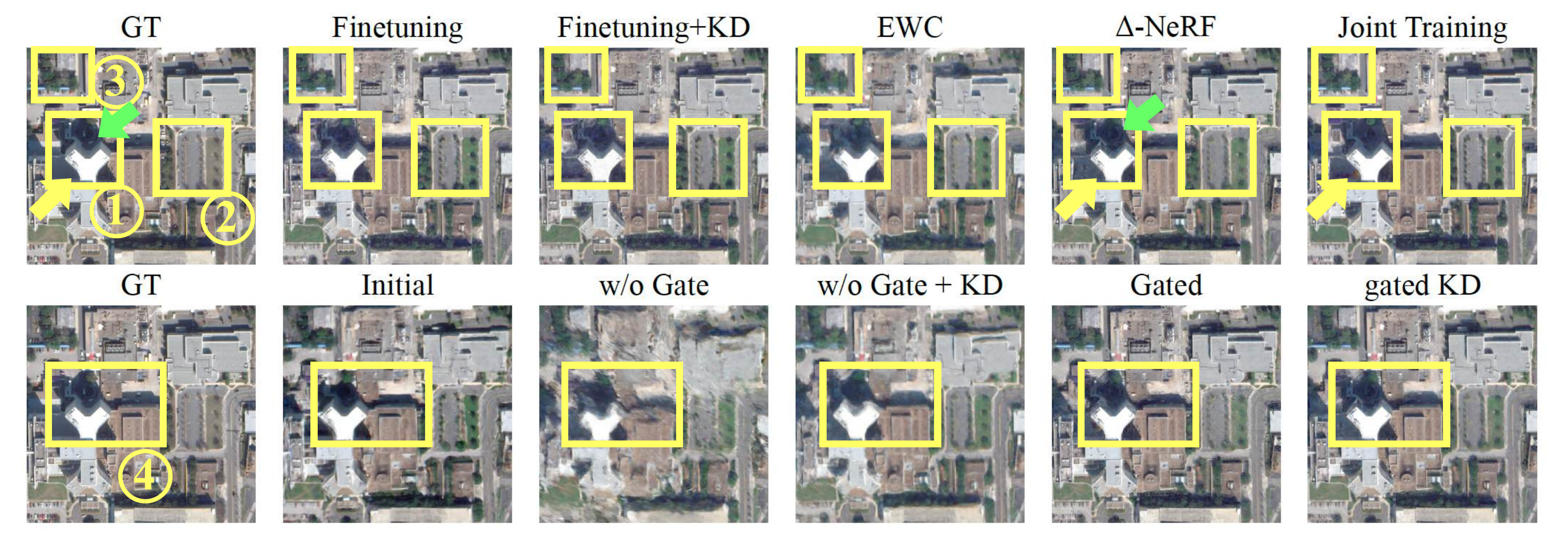}   
    \caption{
Qualitative comparison on an unseen view from JAX-068. The \textit{top row} shows baseline methods vs. ours. The \textit{bottom
row} presents an ablation of our method, comparing variants with and without gating and knowledge distillation. Our full method
(Gated KD) produces the sharpest textures, closely resembling joint training on the full optimized dataset.
}
    \label{fig:qual_comp}
\end{figure*}

\begin{figure*}[t]
    \centering
    \includegraphics[width=\linewidth,height=5.5cm, keepaspectratio]{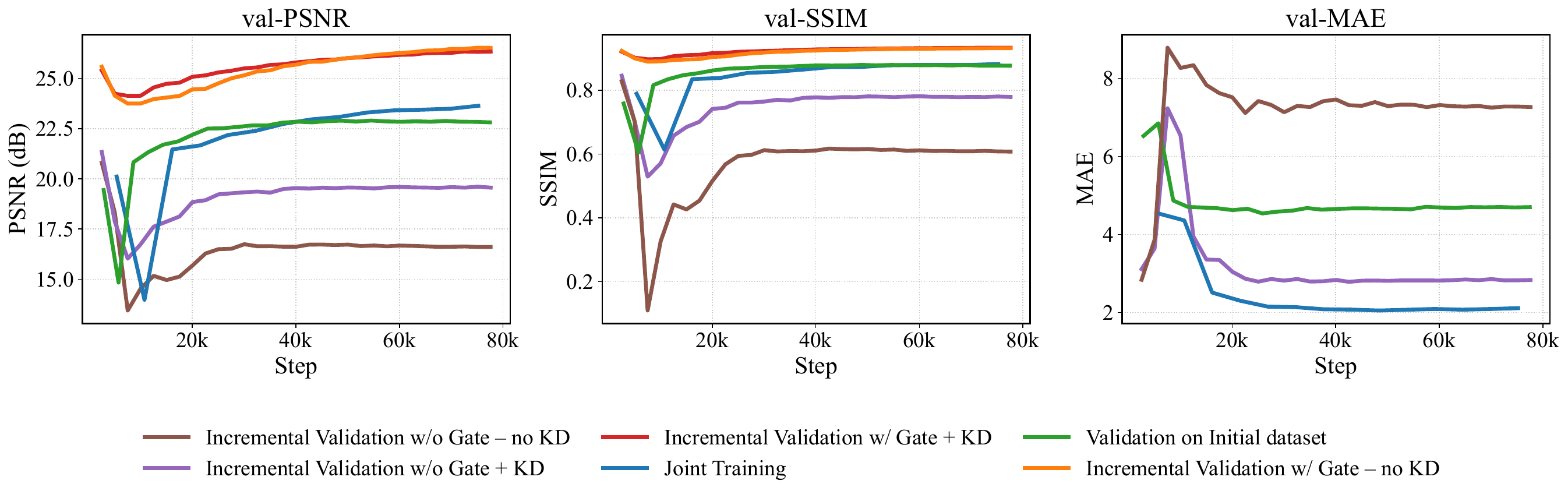}
    \caption{Ablation study on scene JAX-068. Our gated $\Delta$-NeRF with KD steadily improves PSNR and SSIM while reducing MAE. MAE is not gated.}
    \label{fig:ablation}
\end{figure*}

\paragraph{Experimental Setup:}

We evaluate $\Delta$-NeRF with Sat-NeRF as the base model on the DFC2019 Maxar WorldView-3 satellite dataset~\cite{dfc2019}, focusing on four scenes [JAX-004, JAX-068, JAX-214, JAX-260] from Jacksonville, Florida acquired between 2014 and 2016. Each scene comprises $\sim$800$\times$$\sim$800 RGB crops with 0.3m/pixel resolution, covering 256$\times$256 m\textsuperscript{2} area, with RPC metadata, sun angle information, and corresponding digital surface models (DSMs).

\subsubsection{Dataset Processing:}
Each scene contains 9-21 training views and 2-3 test views. We apply our view selection strategy to extract a compact yet diverse set of views. The selected views are split into:
(1) \textbf{Initial dataset}, used for the initial phase of the base model training (2) \textbf{Incremental dataset}, used for training the residual controller incrementally.
All experiments use a shared spatial grid to maintain a consistent normalized coordinate system. We do not use replay or memory buffers; both initial and incremental stages are evaluated on the same unseen test set.

\subsubsection{Training Configuration:}

\begin{table}[t]
\centering
\setlength{\tabcolsep}{1pt}
\begin{tabular}{lp{5.5cm}}
\toprule
\textbf{Method} & \textbf{Description} \\
\midrule
Joint Training     & Trained on the full optimized dataset. \\
Initial Only       & Trained only on initial data; no updates. \\
Finetune           & Naive finetuning on new data. \\
Finetune + KD      & Finetuning with KD from frozen base. \\
EWC                & Incremental finetuning with EWC. \\
\bottomrule
\end{tabular}
\caption{Baseline variants used in our experiments.}
\label{tab:baseline_variants}
\end{table}

We follow the Sat-NeRF training setup as our baseline.
Full-scene training takes $\sim$10 hours (300K iterations), while our optimized view subset converges within $\sim$100k iterations ($\sim$3-5 hours per training stage). The combined initial and incremental training takes $\sim$6-8 hours per scene on an NVIDIA RTX A6000 GPU. Table~\ref{tab:baseline_variants} highlights our baselines.

\subsubsection{Evaluation Metrics:}
The JAX-068 scene serves as our primary example, with full quantitative results across all scenes in Figure~\ref{fig:metrics_barplot} and Appendix I. Evaluation metrics include: PSNR for RGB reconstruction quality, SSIM for perceptual structural similarity and MAE for mean error in altitude for predicted DSMs.

\subsection{Quantitative Results:}
\subsubsection{Across-scene trends:}
To thoroughly evaluate the effectiveness of our residual controller with gated inference in preserving prior knowledge and generalizing to new views, we analyzed PSNR, SSIM, and MAE across all four scenes. As shown in Figure~\ref{fig:metrics_barplot}, our method consistently matches or outperforms all baselines, achieving the best PSNR and SSIM in every case, and even surpassing joint training in scenes like JAX-068 and JAX-260. SSIM improves significantly over finetuning variants, with up to $+53.1\%$ gain (scene JAX-260), while MAE substantially reduces, most notably in scene JAX-260 with a 29.6$\%$ reduction over naive finetuning. These trends confirm the effectiveness of $\Delta$-NeRF even without access to previous data. 
A more detailed presentation of the results can be found in Appendix I.

\begin{table}[t]
\centering
\setlength{\tabcolsep}{4pt} 
\begin{tabular}{lcccc}
\toprule
\textbf{Method} & \textbf{PSNR ↑} & \textbf{SSIM ↑} & \textbf{MAE ↓} & \textbf{Params} \\
\midrule
$\Delta$-NeRF            & 24.31           & 0.891           & 4.097          & 5.2M \\
Compressed    & 22.85           & 0.862           & 4.473          & 1.02M \\
\bottomrule
\end{tabular}
\caption{Performance vs. parameter count for full and compressed models. The compressed version retains most performance while reducing size by $\sim$5×.}
\label{tab:compression}
\end{table}

\subsubsection{Scene-wise example (JAX-068):}
 
We analyze scene JAX-068 as a representative example. Naive finetuning slightly improves MAE but reduces RGB quality due to forgetting (PSNR: 21.43). Our method outperforms all baselines in terms of RGB reconstruction and perceptual similarity. Specifically, our method improves PSNR by 23.5$\%$ over Finetune, 21.6$\%$ over Finetune+KD, and even 9.8$\%$ over Joint Training, demonstrating that residual modulation and gating enhance generalization beyond what joint optimization achieves. SSIM also increases by 9.3$\%$ over Finetune+KD, indicating stronger structural fidelity. In terms of MAE, our approach reduces error by 40.1$\%$ over Initial and 16.6$\%$ over Finetune+KD. Although 0.671m higher than the joint training result, our MAE performance nevertheless confirms its effectiveness. 
\subsection{Qualitative Results:}

Figure~\ref{fig:qual_comp} top row, presents qualitative comparisons of RGB prediction
of various baselines and our approach. The visual reconstruction of joint optimization is very similar to the original baselines. Our incremental approach preserves
sharper edges and more accurate terrain structure, closely
matching joint training and ground truth.
Regions 1-3 highlight key improvements in $\Delta$-NeRF: better structural recovery (e.g., Y-shaped rooftop in region 1), color fidelity (green patch in region 2), texture and shadow sharpness (trees in region 3), and fine detail reconstruction (the roundabout marked by a green arrow, visible only in GT and $\Delta$-NeRF). A red artifact introduced by joint training but absent in GT or $\Delta$-NeRF is marked with a yellow arrow.

Figure~\ref{fig:qual_rgb_dsm}  shows DSM predictions. Our method significantly improves upon the noisy initial DSM, producing results comparable to those of joint optimization, albeit with some edge blurring in incremental variants.

\subsection{Optimization Result:}

Our view selection strategy reduces the JAX-068 training set from 17 to 9 images (47\% reduction), cutting training time from 10 to $\sim$4 hours (60\% reduction) in joint training while maintaining comparable appearance quality. This efficiency comes with a modest MAE trade-off (+0.885 from original). The incremental training time is reduced  by 30-42\% on a NVIDIA RTX A6000 GPU (Table~\ref{tab:training_time2}).

\begin{table} 
\centering
\setlength{\tabcolsep}{2pt} 
\begin{tabular}{lcccc|c|c}
\toprule
 & \multicolumn{2}{c}{Initial} & \multicolumn{2}{c|}{Incr.} & Total & Time Est. \\
\cmidrule(lr){2-3} \cmidrule(lr){4-5}
Method & Raw & Opt & Raw & Opt & Views & (hrs.) \\
\midrule
Retrain & 7 & \ding{55} & 10 & \ding{55} & 17 & 4 + 10 = 14  \\
$\Delta$-NeRF       & 7 & 5 & 10 & 4 & 9 & 3 + 5 = 8 \\
\bottomrule
\end{tabular}
\caption{Retraining vs $\Delta$-NeRF training time on JAX-068}
\label{tab:training_time2}
\end{table}

Post-training, we apply KD to compress the $\Delta$-NeRF ensemble from 5.2M to 1.02M parameters (Table~\ref{tab:compression}). Despite this 80$\%$ reduction, the compressed student achieves strong performance: PSNR drops modestly by $\sim$1.46 dB, SSIM by $\sim$0.029, and MAE increases by only $\sim$0.376 for JAX-260. 

\subsection{Ablations and Insights:}

Figure~\ref{fig:ablation} analyzes the effect of each component:
\begin{itemize}
\item \textbf{Controller vs. No Controller:} Removing the residual controller during incremental refinement causes a sharp performance drop, highlighting its role in adapting to new views.
\item \textbf{KD vs. No KD:} Knowledge distillation adds regularization to the residuals and yields modest gains, though controller-based gating alone performs comparably to strong baselines.
\item \textbf{Gated Inference:} Uncertainty-aware gating substantially improves RGB quality preventing overcorrection. In the example scene, gating boosts PSNR from 19.53 to 26.46—a $35.5\%$ improvement—though MAE gains primarily result from controller trunk modulation.
\end{itemize}

Figure~\ref{fig:qual_comp} (bottom row) shows qualitative ablations. Region 4 highlights the structural clarity of our Gated KD variant, which preserves sharp edges and avoids the smudging seen in other variants.

\paragraph{Insights and Analyses:}
(1) In some scenes (e.g., JAX-260), our method even surpasses joint optimization in PSNR and SSIM — likely due to residual regularization, gated correction, and the use of optimized, informative views, which reduce overfitting and enhance generalization.

(2) Without gating, the controller over-corrects despite the base being frozen, simulating catastrophic forgetting. Gating ensures seamless integration of old and new knowledge and prevents such over-correction. 

(3) Without gating, RGB performance shows little improvement over initial training, however MAE still improves directly from controller modulation. 

(4) Working with optimized datasets, retaining maximal information is crucial. While well-selected views approximate full-data performance, suboptimal ones degrade results significantly—especially MAE—underscoring the value of depth-aware selection. 

(5) Although our results focus on urban scenes, the method is scalable to natural terrains due to joint depth–appearance based view selection. However, generalization to diverse environments depends on enhancing the base model itself.

Overall, our approach avoids forgetting, reduces training cost, and requires no access to prior data while matching joint training.

\section{Conclusion}
We propose a novel framework for incremental refinement of NeRFs on satellite imagery. Our $\Delta$-NeRF refines a frozen base model without forgetting, giving comparable performances with joint training and other baselines under limited data. Unlike prior work requiring replay or pose estimation etc., our method enables uncertainty-aware gating and optimized view selection. Future extensions include using the controller for explicit change detection.

\bibliography{aaai2026}

\end{document}